\documentclass{article} 
\usepackage[final]{colm2026_conference}

\usepackage{microtype}
\usepackage{hyperref}
\usepackage{url}
\usepackage{booktabs}
\usepackage{graphicx}
\usepackage{caption}
\usepackage{subcaption}
\usepackage{wrapfig}
\usepackage{longtable}
\usepackage{colortbl}
\usepackage[most]{tcolorbox}
\newtcolorbox{promptbox}{colback=gray!7,colframe=gray!55,boxrule=0.4pt,left=5pt,right=5pt,top=3pt,bottom=3pt,fontupper=\small\itshape,arc=2pt}
\usepackage{amsmath}
\usepackage{lineno}

\definecolor{darkblue}{rgb}{0, 0, 0.5}
\hypersetup{colorlinks=true, citecolor=darkblue, linkcolor=darkblue, urlcolor=darkblue}

\title{Where did the ambiguity go? Examining how\\ multimodal models interpret polysemous words}

\author{Jasin Cekinmez\thanks{Equal contribution.}\quad
Addison J.~Wu\footnotemark[1]\quad
Raja Marjieh\quad
Thomas L.~Griffiths \\
Princeton University \\
\texttt{\{jasincekinmez, addisonwu\}@princeton.edu} \\
Code: \url{https://github.com/addisonwu05/llm-polysemy}
}

\begin{document}

\ifcolmsubmission
\linenumbers
\fi

\maketitle

\begin{abstract}
Human language is highly polysemous. Many common words (e.g., ``bank'' or ``palm'') carry several distinct meanings that shape what humans communicate and imagine. Large language models (LLMs) have been shown to understand this multiplicity of meaning, but much less is known about how polysemy surfaces in other modalities such as images. We study this across 17 text-to-image and 15 text-generation models by giving each a polysemous word with no context to fix its meaning and measuring which senses are produced over many samples. We find a clear multimodal gap, where within every model family, generated images settle on far fewer senses than generated sentences (normalized entropy $0.10$ vs.\ $0.25$), and both are far less varied than what people imagine for the same words (normalized entropy $0.47$). However, when we instead ask a model to list how often it would generate outputs corresponding to each possible meaning of a word, it predicts distributions that are more diverse than the actual space of outputs. These results reveal a multimodal gap in how foundation models express meaning, and how their understanding may not transfer faithfully nor equally across modalities.
\end{abstract}

\section{Introduction}

Large language models \citep{brown2020language,achiam2023gpt,touvron2023llama} and their multimodal successors \citep{team2023gemini,alayrac2022flamingo,liu2023visual,team2024chameleon}, have moved from being research artifacts to everyday tools. These systems can draft prose, answer questions, and synthesize photorealistic imagery on demand \citep{saharia2022photorealistic}, being used at a scale large enough to influence what people read, see, and imagine \citep{bommasani2021opportunities,eloundou2024gpts}. Thus, it matters not just whether any single output is good, but whether the \emph{distribution} of outputs a model produces looks like the behavior of the system in aggregate \citep{rahwan2019machine}.

The properties of this distribution are a growing concern. Generative models are often observed to homogenize, converging on a narrow, repetitive band of responses and flattening the variety present in their training data and in human expression \citep{bender2021dangers,weidinger2021ethical,anderson2024homogenization, shumailov2024ai}. Such collapse has implications beyond any one interaction, raising the prospect of a feedback loop in which machine-shaped culture narrows the space of ideas people are exposed to \citep{brinkmann2023machine,messeri2024artificial,glickman2025human}. Yet this possibility has been studied almost entirely in open-ended \emph{text} generation. Far less is known about whether the same narrowing occurs in the \emph{visual} outputs of multimodal models, how it differs across modalities, and how either compares to the variability of human responses.

\begin{figure}[t!]
    \centering
    \includegraphics[width=\linewidth]{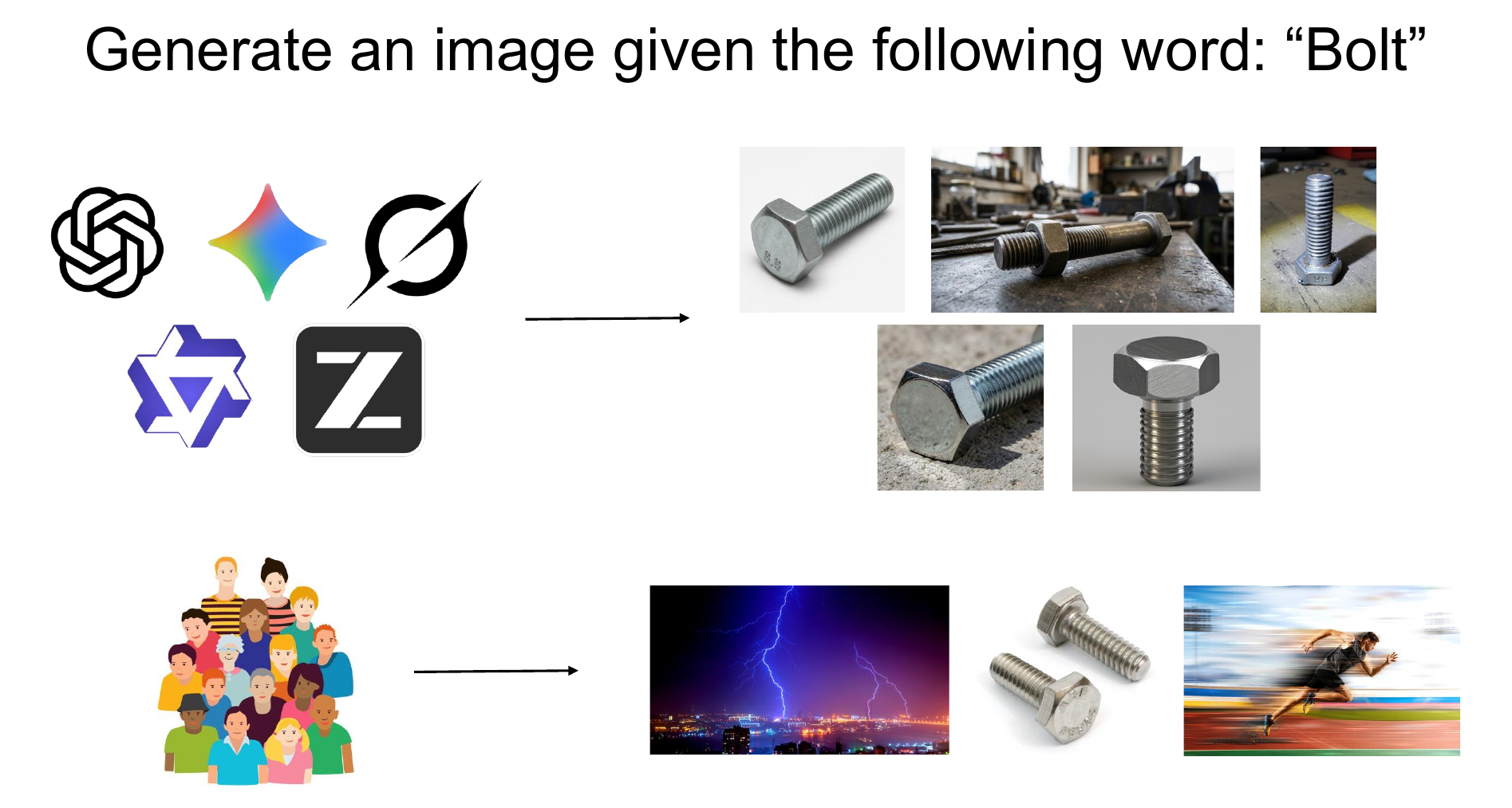}
    \caption{\textbf{We give the same context-free polysemous word to text-to-image models (top) and to people (bottom), and compare which senses each produces.} For ``bolt'', every model depicts a single sense, a metal fastener, whereas people split across the word's distinct meanings: lightning, a fastener, and sprinting. Text-to-image models collapse a polysemous word onto one dominant sense while humans preserve its ambiguity.}
    \label{fig:teaser}
\end{figure}

We argue that \emph{polysemy} offers a clean probe for addressing these questions.\footnote{Strictly, lexical semantics distinguishes \emph{polysemy}, in which one word carries multiple related senses (the \textit{mouth} of a person versus of a river), from \emph{homonymy}, in which unrelated meanings happen to share a form (a river \textit{bank} versus a financial one). By this criterion several of our examples, \textit{bank} among them, are in fact homonyms. Because our concern is simply the distribution of meanings a model commits to for a word presented in isolation, this distinction does not bear on our analysis. We use \emph{polysemy} as an umbrella term for words with multiple meanings.} Human language is pervasively ambiguous \citep{miller1995wordnet}, as many common words carry several distinct senses (\textit{bank}, \textit{palm}, \textit{bolt}; Figure~\ref{fig:teaser}). This multiplicity is argued to be a functional feature of efficient communication rather than a defect \citep{piantadosi2012communicative}.  Within a context, the intended sense is disambiguated. Without it, such a word is genuinely ambiguous, so any commitment to a subset of its meanings reveals a model's prior. Presenting the same word to a model and a human and asking which sense surfaces turns a linguistic phenomenon into a measurable distribution over meanings, and asking it across modalities tests whether meaning transfers faithfully from language to vision.

We measure this with 100 polysemous words, each with a finite set of senses, presented to many text-to-image and text-generation models across multiple families. For every word, we draw many independent samples and classify each output's sense, yielding, for each model and word, a distribution over senses. The image and text conditions share the same words, sense inventories, and judges, so their distributions are directly comparable. Moreover, we anchor both against a human baseline collected under matched framings.

Across all models we find a pronounced \emph{multimodal gap}. Image models commit to a far narrower set of senses than text models within the same family, and both are markedly less diverse than humans. Notably, the models are not unaware of ambiguity. When asked to predict the distribution of senses, they estimate human-like diversity, far above what they themselves generate. Preference-tuning a generator can sharpen it further, though alignment alone does not explain the full gap. Together, these results expose a gap between what multimodal models know about meaning and what they reveal when they produce it, with direct consequences for how faithfully understanding transfers across modalities.

\section{Related Work}

\subsection{Semantic Representation in Foundation Models}

The representation of word meaning, or semantic representation, is a classic problem in machine learning and cognitive science \citep{miller1995wordnet,griffiths2007topics,landauer1997solution,rogers2004semantic,mikolov2013word2vec}, with polysemous words posing a particular challenge. Static word embeddings \citep{mikolov2013word2vec,pennington2014glove} potentially force a single vector to carry every sense of a polysemous word. \citet{arora2018linear} showed that such an embedding is a linear superposition of sense vectors weighted by corpus frequency. Contextual encoders \citep{devlin2019bert} move one step further by assigning words context-dependent vectors which recover rich semantic structure \citep{rogers2020primer}, although their geometry is only partly sense-disentangled \citep{ethayarajh2019contextual}. Superposition also arises in mechanistic interpretability as neural networks represent more features than dimensions \citep{elhage2022superposition,templeton2024scaling}. Such representational pressures carry into multimodal models: CLIP \citep{radford2021clip} conditions most text-to-image systems, and \citet{white2022schrodinger} showed how it encodes a homonym as a superposition of its senses. Thus, the downstream generator must resolve a representation that is ambiguous by construction.

\subsection{Ambiguity in Generative Models}

Lexical ambiguity is long-studied in natural language processing, from the existence of discrete senses \citep{kilgarriff1997dont} to word-sense disambiguation \citep{navigli2009word}. In context, models can resolve it readily; without context (as in our setting), ambiguity remains unresolved and the model falls back on a prior. Recent work shows LLMs struggle to model ambiguity faithfully \citep{liu2023were,tamkin2023task}, but almost exclusively in \emph{text}. Images make ambiguity unavoidable, because a renderer must commit to a meaning even for equivocal sentences.

Text-to-image generation moved from diffusion \citep{ho2020denoising,rombach2022high}, like with DALL-E 2 \citep{ramesh2022hierarchical}, to auto-regressive systems, like GPT Image \citep{openai2025gptimage1}. A parallel literature documents failures on the compositional and referential structure ambiguity stresses: spatial relations \citep{conwell2022testing}, attribute binding \citep{rassin2023linguistic}, counting and negation \citep{marcus2022preliminary}, and sensitivity to under-specified prompts \citep{hutchinson2022underspecification}, now probed at scale \citep{huang2023t2i}. Most directly, \citet{rassin2022dalle} show DALL-E 2 violates the single-meaning-per-symbol principle, rendering a homonym like \textit{bat} as two entities at once. \citet{kaskov2025undoubling} formalize this as \emph{homonym duplication}, measuring it across diffusion models and identifying an Anglocentric amplification via translation. These works establish that image models can render multiple senses \emph{simultaneously}. We ask the complementary distributional question, across many samples and matched text and human baselines, \emph{which} sense a model selects. We find the dominant failure is not duplication but an implicit narrowing of the sense distribution.

\subsection{Diversity and Homogenization in Foundation Models}

Language models have recently been shown to exhibit mode collapse. \citet{jiang2025artificial} document an ``Artificial Hivemind'' of intra-model repetition and inter-model convergence, warning that ensembles sharing training priors do not recover diversity. Mode collapse is visible at decoding \citep{holtzman2020curious} and amplified by alignment which sharpens models toward high-reward modes at the cost of diversity \citep{ouyang2022training,kirk2024understanding}. At a collective level, shared foundation models predict correlated behavior \citep{bommasani2022picking}, recursive training on generated data erodes distributional tails \citep{shumailov2024ai}, and using the same AI assistant reduces the diversity of \emph{people's} outputs \citep{padmakumar2024does,doshi2024generative,anderson2024homogenization}.

Homogenization is also tied to bias. Models encode and amplify stereotypes \citep{caliskan2017semantics,bolukbasi2016man,nadeem2021stereoset}, and text-to-image systems sharpen demographic skew beyond their training data \citep{bianchi2023easily,luccioni2023stable,cho2023dall}. \citet{wu2026large} show LLMs can develop \emph{novel} biases through insufficient exploration, with newer models exercising more segregative allocation over time.
Multilingual systems stay English-centric in their knowledge \citep{cekinmez2025adam,wendler2024llamas,veselovsky2025localized}. Such stable preferences make models identifiable, with generated images carrying model-specific signatures \citep{yu2019attributing}, allowing \citet{cekinmez2026guess} to recover a model's identity from its images with near-perfect accuracy.

\section{Methodology}

\subsection{Experimental Setup}
Our stimuli are 100 polysemous words, each with a human-verified, closed inventory of candidate senses (the full list, for all three languages considered, is in Appendix~\ref{app:words}), presented in isolation. The prompt is the bare word with no disambiguating context. This forces each model to commit to a sense from its own priors, making the distribution of senses across repeated samples the unit of analysis.

The same word set is presented in two modalities. In the image condition, a text-to-image model receives the bare word as its prompt and renders an image, and we ask which sense the image depicts. In the text condition, an autoregressive language model receives the instruction "Use the following word in a single sentence," and we ask which sense the resulting sentence uses. We evaluate a panel of 17 text-to-image and 15 text generation models spanning the major model families, each tagged with its architecture (diffusion vs. autoregressive) so that behavior can later be related to architectural class. For every word-model pair we draw 30 independent samples.

Because the image and text conditions share the same words and the same closed sense inventories, their resulting sense distributions are directly comparable. A difference between the two modalities reflects how the models resolve the word, not a difference in what was asked or how it was scored.

\subsection{Models}
Our central panel spans the major providers in both modalities: OpenAI (\texttt{gpt-image} and GPT,~\citealp{achiam2023gpt}), xAI (Grok), Google (Gemini,~\citealp{team2023gemini}), Alibaba (Qwen), and Z.ai (GLM). Each provider contributes several recent generations, with the exact checkpoints shown in the figures; the complete inventory, with references, is in Appendix~\ref{app:models}. Beyond this panel,  targeted analyses add open comparison models with recorded: the architecture analysis contrasts diffusion models, FLUX, and autoregressive generators, and the preference-tuning ablation pairs SDXL~\citep{podell2023sdxl} and SimpleAR~\citep{wang2025simplear} with their aligned checkpoints, holding the base model fixed.

\subsection{Sense Classification}

Using GPT 5.4 as our judge, each image/text generation is assigned to one of the word's senses or a fallback label: multiple ($>2$ senses at once), unclear, or other (prompts in Appendix~\ref{app:prompts}). A multiple verdict triggers a follow-up listing the senses seen, so superposition is quantified rather than discarded. For each word-model cell we collapse the 30 labels into a distribution over senses. This per-cell distribution is then analyzed as we described in Section 3.3. To confirm the judge labels reliably, an author hand-labeled a random sample of 60 English images blind to the judge's verdict and agreed with it on all 60.

We also collect a human reference baseline from Prolific over the same 100 words (Appendix~\ref{app:human}), scored by the same judge, under two framings matched to the modalities: ``use the following word in a sentence'' (meaning vs. text models) and ``what image does this word evoke'' (image vs. image models).

\subsection{Distributional Metrics}

We summarize each model's behavior on a word as a sense distribution: the fraction of its 30 samples that the judge assigned to each candidate sense, with `multiple', `other', and `unclear' as additional categories. These fallback labels/additional categories are rare ($6.4\%$ of all samples; $8.0\%$ of image and $4.5\%$ of text outputs), so they contribute little to the estimates below (per-model and human rates in Appendix~\ref{app:fallback}). All of our analyses compare these distributions using three standard quantities which we describe next.

\subsubsection{Jensen--Shannon Similarity}  
To measure how differently two models resolve a word, we use the Jensen--Shannon (JS) similarity between their sense distributions $p$ and $q$,

\[
JSS(p,q)
=
1-\sqrt{
\frac{1}{2}\mathrm{KL}(p\|m)
+
\frac{1}{2}\mathrm{KL}(q\|m)
},
\qquad
m=\frac{1}{2}(p+q).
\]

Here $\mathrm{KL}$ denotes the Kullback--Leibler divergence and $m$ is the mean distribution. Using base-2 logarithms, $JSS$ is bounded in $[0,1]$. It equals $1$ when the distributions are identical and $0$ when they have disjoint support. We use JS similarity throughout the paper to construct model-similarity matrices and hierarchical clustering, quantify within- and between-family cohesion, measure per-word inter-model disagreement, compare image and text models, and assess each model's distance from corpus-frequency priors.

\subsubsection{Shannon entropy} 
To measure how decisively a model commits to a single sense, we use the Shannon entropy of its sense distribution. For a distribution $p$ over the senses $s \in V$, entropy is
$$
H(p) = -\sum_{s \in V} p_s \log_2 p_s,
$$
measured in bits. It quantifies the uncertainty of the distribution: $H=0$ when all of the model's samples fall on a single sense (a fully decisive model), and $H$ grows as the samples spread more evenly across senses. For a word with $|V|$ senses the maximum is $\log_2 |V|$ bits, attained by the uniform distribution. Because words differ in how many senses they have, we normalize by this maximum to obtain the normalized entropy $\tilde{H}$ defined as
$$
\tilde{H}(p) = \frac{H(p)}{\log_2 |V|} \in [0,1],
$$
so that $0$ marks a model that always picks the same sense and $1$ a model that is maximally diverse, on a scale comparable across words. We use $\tilde{H}$ to compare sense diversity across modalities, against the human baseline, and across model generations.

\subsubsection{Principal component analysis}
To visualize model behavior, we represent each model as the concatenation of its per-word sense distributions into a single feature vector, standardize the features, and project to two dimensions with PCA. Models that resolve polysemy similarly appear close together. We apply this within each modality and to the pooled image-and-text set to ask whether the dominant axis of variation reflects modality or model family.
\clearpage
\section{Results}
\begin{figure}[t]
    \centering
    \includegraphics[width=0.85\linewidth]{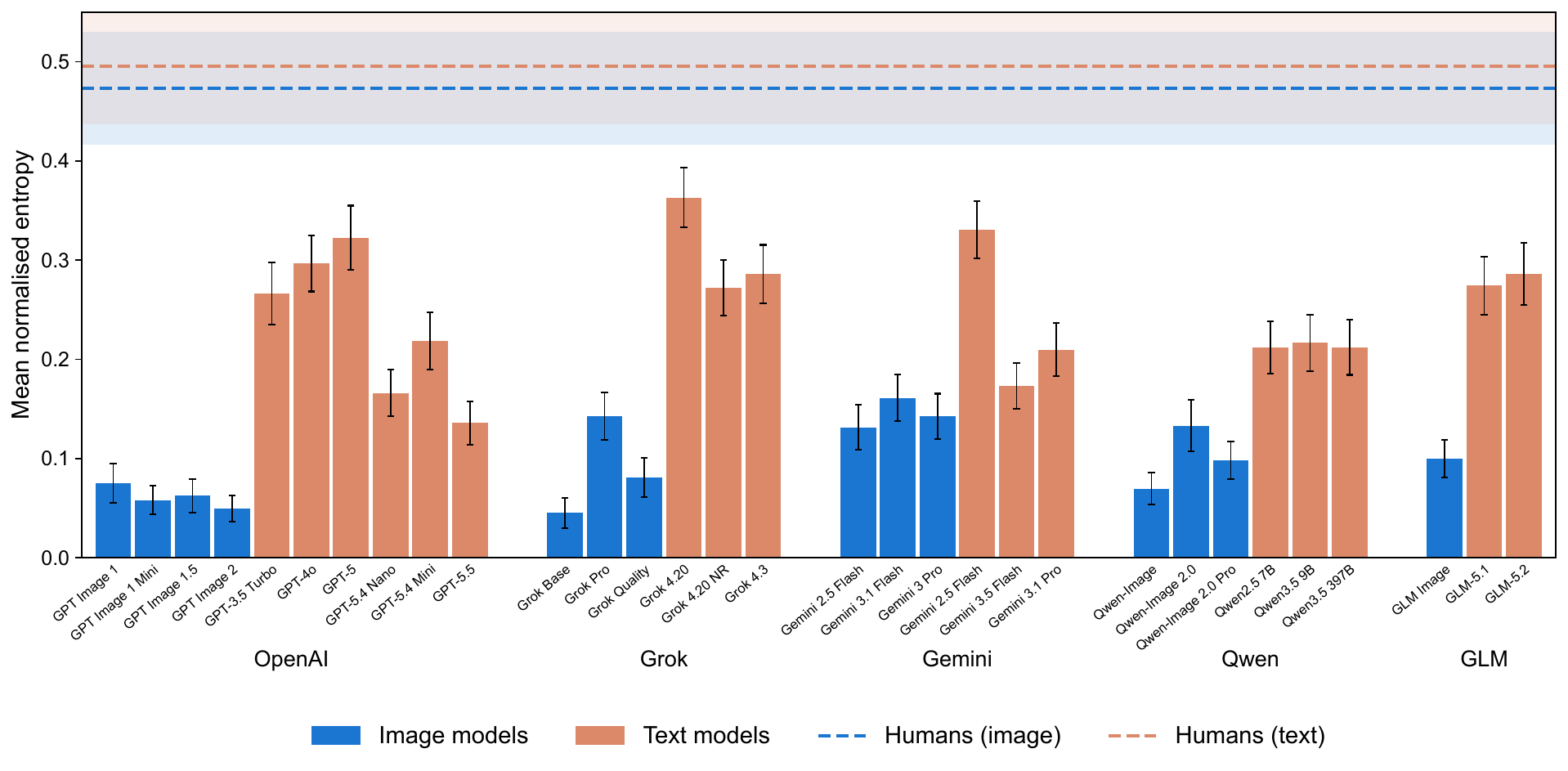}
    \caption{Sense diversity across models and modalities, against the human baseline. Generated images commit to far fewer senses than generated text in nearly every family, and both modalities fall well short of human diversity.}
    \label{fig:modality_gap}
\end{figure}
\subsection{The Multimodal Gap}

\textbf{Image models are less semantically diverse than their counterpart text models within every provider family}. As shown in Figure \ref{fig:modality_gap}, the semantics of the images generated by image-generation models in each family have consistently lower normalized entropies in their semantic diversity, compared to what is observed in the sentences generated by the text models. On average, image models reach a normalized entropy of only $0.10$, compared to $0.25$ for text models (paired Wilcoxon $p < 10^{-6}$). The gap is consistent across families, ranging from $0.09$ (Gemini) to $0.22$ (Grok). It is also not an artifact of the words' grammatical class: splitting the panel into the $50$ words that carry a salient verb, adjective, or pronoun sense and the $50$ that are noun-dominant, the two halves are statistically indistinguishable in diversity within each modality, and the image--text gap persists within each half (Appendix~\ref{app:pos}).

\textbf{Image and text-generation models consistently display less semantic diversity than humans.} When human participants were asked what image a polysemous word evokes, people reach a mean normalized entropy of $0.47$ (Figure~\ref{fig:modality_gap}), far above every AI model. Crucially, humans show no modality gap of their own: they are about equally diverse whether imagining a word ($H_\textrm{norm} = 0.47$) or using it in a sentence ($H_\textrm{norm} = 0.50$; paired $t$-test $p = 0.31$). The image-versus-text collapse is therefore specific to models, where it is large ($d = 0.80$), rather than a property of how the two modalities are naturally used.

\begin{wraptable}{r}{0.60\linewidth}
\centering
\small
\setlength{\tabcolsep}{4.5pt}
\vspace{-\baselineskip}
\begin{tabular}{lcccccc}
\toprule
 & \multicolumn{2}{c}{English} & \multicolumn{2}{c}{Turkish} & \multicolumn{2}{c}{French} \\
\cmidrule(lr){2-3}\cmidrule(lr){4-5}\cmidrule(lr){6-7}
Family & Text & Img & Text & Img & Text & Img \\
\midrule
OpenAI & 0.234 & 0.061 & 0.287 & 0.148 & 0.217 & 0.072 \\
Grok   & 0.307 & 0.090 & 0.274 & 0.169 & 0.231 & 0.099 \\
Gemini & 0.238 & 0.145 & 0.312 & 0.239 & 0.221 & 0.134 \\
\midrule
\textit{All} & 0.253 & 0.095 & 0.290 & 0.182 & 0.222 & 0.099 \\
\bottomrule
\end{tabular}
\caption{Cross-lingual modality gap ($H_{\mathrm{norm}}$). Image $<$ text in every language; the gap is smallest in Turkish.}
\label{tab:language_entropy}
\end{wraptable}
\textbf{Low diversity is observed when we conduct the same experiment in other languages}. To assess whether our findings generalize across languages, we considered 25 polysemous words in both Turkish and French. We find that the tendency of models to exhibit reduced sense diversity is largely language-invariant: averaging over the OpenAI, Grok, and Gemini families evaluated in all three languages, text models reach $H_\textrm{norm} = 0.25$, $0.29$, and $0.22$ in English, Turkish, and French, while image models reach $0.10$, $0.18$, and $0.10$ respectively (Table~\ref{tab:language_entropy}). The modality gap holds in every language, $0.16$ (English), $0.11$ (Turkish), and $0.12$ (French).

\begin{wraptable}{r}{0.40\linewidth}
\centering
\small
\vspace{-\baselineskip}
\begin{tabular}{lccc}
\toprule
Framing & Pred. & Gen. & Human \\
\midrule
Meaning & 0.721 & 0.246 & 0.496 \\
Image   & 0.730 & 0.103 & 0.473 \\
\bottomrule
\end{tabular}
\caption{Predicted vs.\ generated vs.\ human ($H_{\mathrm{norm}}$). Models overestimate human diversity yet generate little diversity themselves.}
\label{tab:stated_revealed}
\end{wraptable}\textbf{Nevertheless, large language models are still aware of semantic ambiguity, even if it is not strongly reflected in their generated samples}. We asked the text-generating models within each family to predict the distribution of interpretations that people would assign to each polysemous word. Models consistently predicted substantially more diverse sense distributions ($H_{\mathrm{norm}}\approx0.73$) than those observed in human judgments ($H_{\mathrm{norm}} \approx 0.47$), suggesting that they overestimate the prevalence of less common senses. This overestimation is even more pronounced when compared to the models' own generations ($H_{\mathrm{norm}}=0.246$ for language and $0.103$ for image generation), revealing a large gap between predicted and revealed semantic preferences~(Table~\ref{tab:stated_revealed}). This overestimation is one of calibration rather than confusion about which senses matter: the predicted distributions track the human sense ordering closely (median Spearman $\rho = 0.87$; the human-dominant sense is the model's top prediction $72\%$ of the time, versus $32\%$ at chance), but systematically flatten it, shifting probability mass off the dominant sense (from $0.72$ down to $0.55$) onto rarer ones rather than reordering which senses matter (Appendix~\ref{app:calibration}).

\subsection{Comparing Sense Selection Across Models}

\begin{figure}[t]
    \centering
    \includegraphics[width=0.85\linewidth]{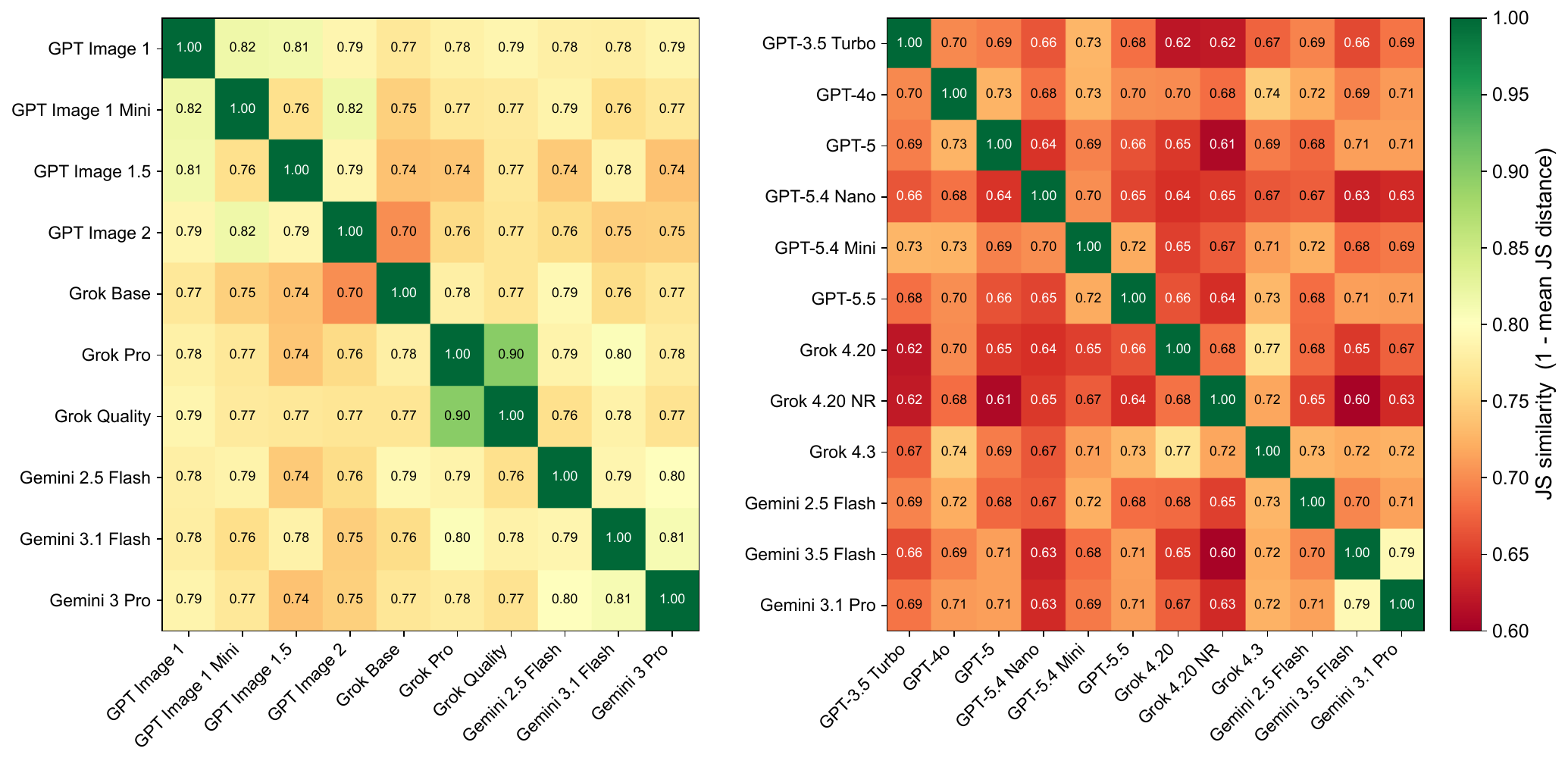}
    \caption{Pairwise sense-distribution similarity between models in each modality (left: image; right: text). Different image generators (left) converge on the same sense more than different text models (right) do (mean pairwise JS-similarities $0.78$ vs.\ $0.69$).}
    \label{fig:similarity}
\end{figure}

\textbf{Different image models converge on the same sense more than different text models do.} Given a fixed modality, for all OpenAI, Grok, and Gemini models, we compute the pairwise JS-similarities of the semantic output distributions between every pair of models, and average over all words (Figure~\ref{fig:similarity}). Image models are more alike than text models, with mean pairwise JS-similarities of $0.78$ compared to $0.69$, respectively ($p < 10^{-3}$).

\begin{figure}[t]
    \centering
    \includegraphics[width=0.9\linewidth]{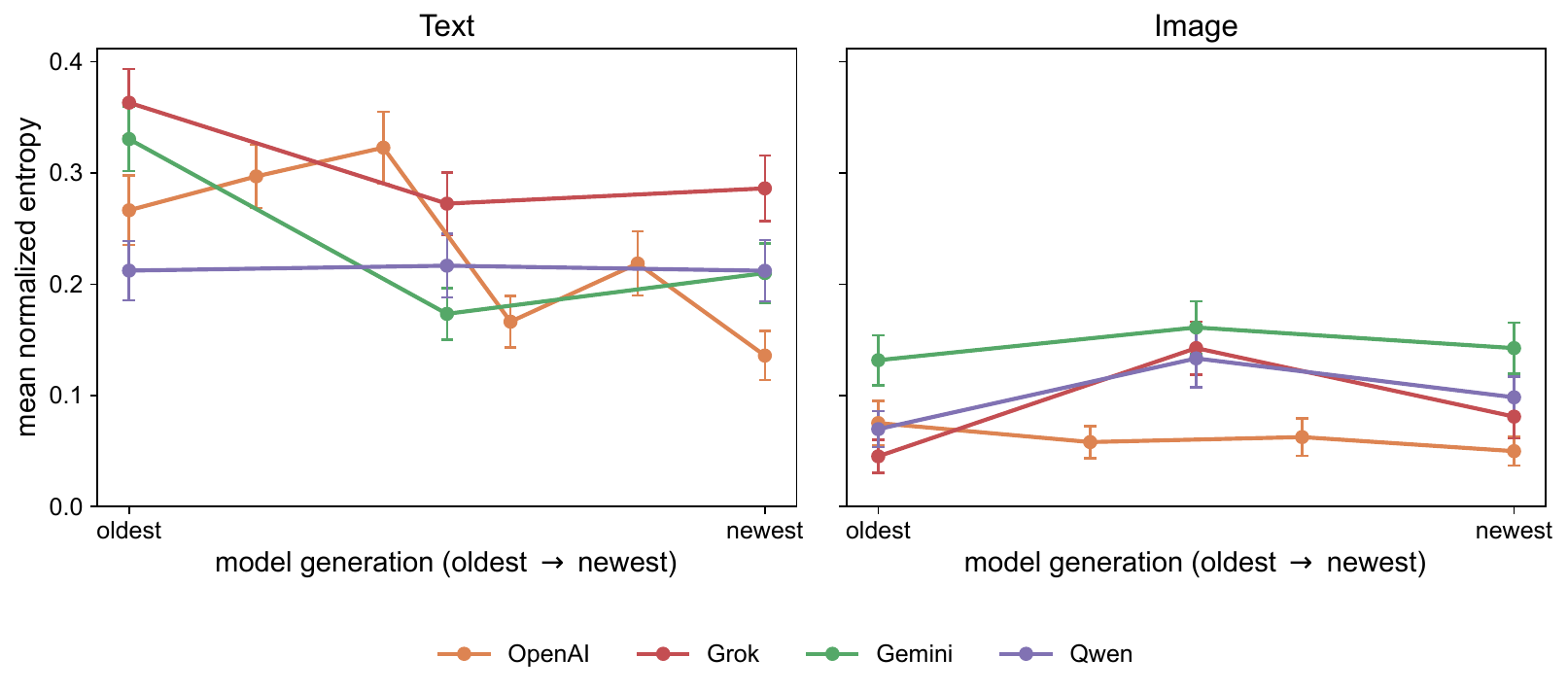}
    \caption{Sense diversity from the oldest to the newest model in each family (left: text; right: image). Text models become steadily less diverse with each generation (Spearman $\rho = -0.59$, $p = 0.02$); image models show no trend and stay far lower. Error bars are $\pm 1$ standard error of the mean across the $100$ words.}
    \label{fig:model_lineage}
\end{figure}

\textbf{Diversity in textual outputs decreases with newer models, while semantic diversity in visual outputs remains low throughout.} Plotting each model's mean normalized entropy against its generation index (Figure~\ref{fig:model_lineage}), textual diversity falls steadily with newer models (Spearman $\rho = -0.59$, $p = 0.02$), and within every text family the newest model is the least diverse (e.g.,\ OpenAI's GPT-3.5 Turbo $\rightarrow$ GPT-5.5, $0.27 \rightarrow 0.14$; Gemini's 2.5 Flash $\rightarrow$ 3.1 Pro, $0.33 \rightarrow 0.21$). Image models show no such trend ($\rho = +0.19$, $p = 0.53$) but sit far below text at every generation ($H_\textrm{norm} \approx 0.10$ vs.\ $0.25$), leaving little room to collapse further.

\begin{figure}[t]
    \centering
    \includegraphics[width=0.85\linewidth]{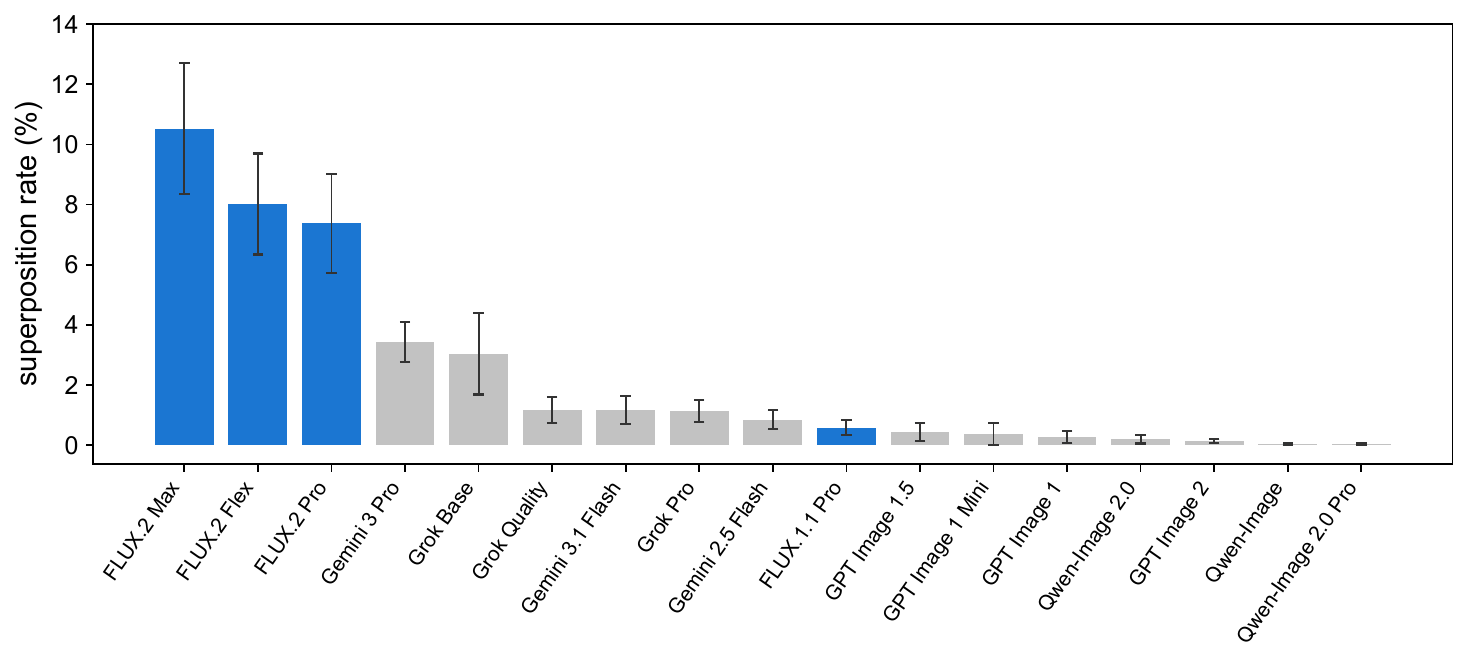}
    \caption{Per-model rate of generated images that blend multiple senses into one. The diffusion-based FLUX models (blue) superimpose senses far more than any other image model ($6.6\%$ vs.\ $0.9\%$; model-level $p = 0.01$). Error bars are $\pm 1$ clustered standard error from a word-level bootstrap ($2000$ resamples of the $100$ words), which accounts for superposition being strongly word-dependent.}
    \label{fig:superposition}
\end{figure}

\textbf{Diffusion image models uniquely superimpose multiple senses into a single image.} The two LLM judges flag a generated image as depicting two or more senses at once for $6.6\%$ of outputs from the diffusion-based FLUX models (Figure~\ref{fig:superposition}), versus just $0.9\%$ for every other image family (model-level Mann--Whitney $p = 0.01$). The effect is concentrated in the FLUX.2 variants ($7$--$11\%$). No other model in the panel exceeds $3.4\%$. Rather than committing to a single reading of the word, these diffusion models blend its competing senses into one composite image.

\begin{figure}
    \centering
    \includegraphics[width=0.75\linewidth]{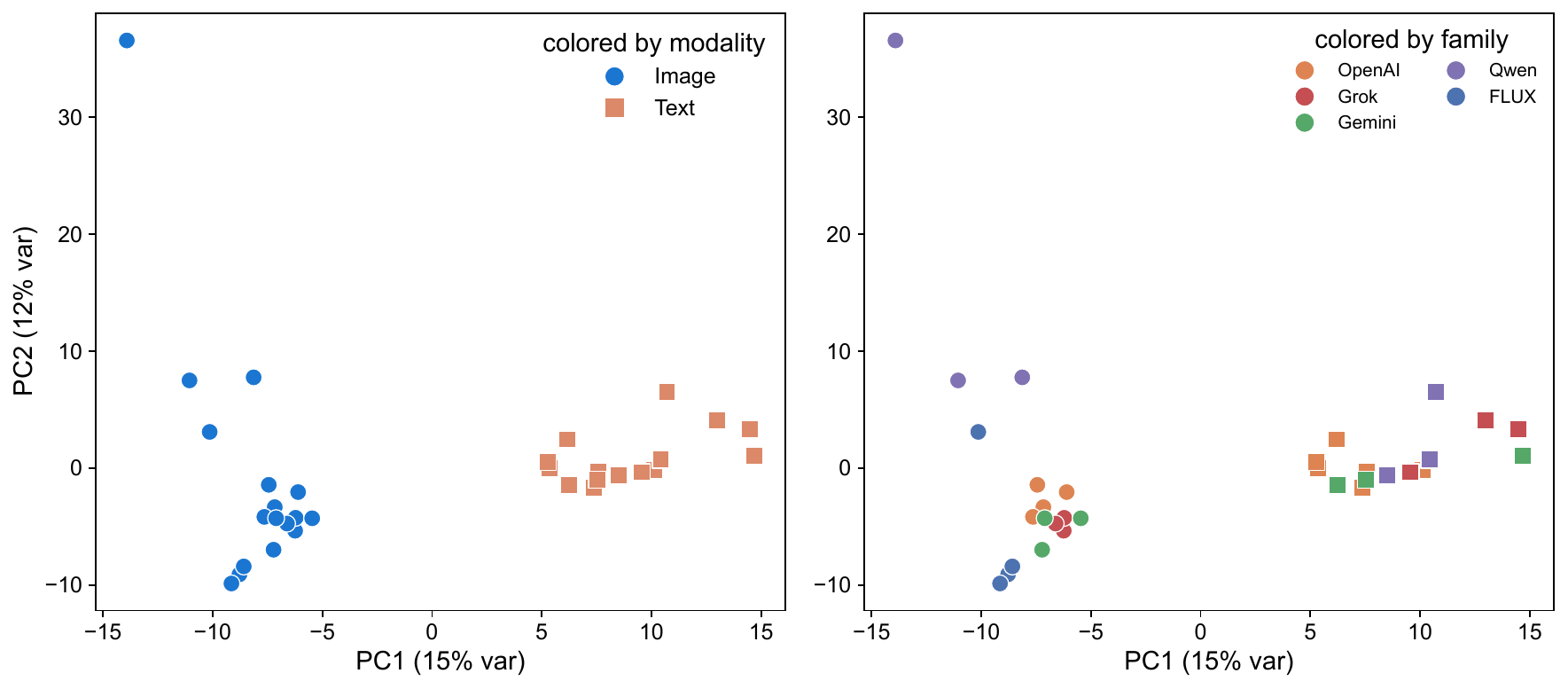}
    \caption{All models embedded in one sense-distribution space (PCA; left colored by modality, right by provider family; circles are image models, squares text). Models split by modality along PC1, not by provider: silhouette $+0.27$ by modality versus $-0.04$ by family.}
    \label{fig:pca}
\end{figure}

\textbf{Modality, not model family, is the primary axis of how models resolve ambiguity.} Embedding every model in one sense-distribution space and projecting to two dimensions (Figure~\ref{fig:pca}), the models split along the first principal component by \emph{modality}: every image generator falls on one side and every text model on the other, regardless of provider. A silhouette analysis confirms it: clustering by modality scores $+0.27$, whereas clustering by provider scores $-0.04$ (no structure). An image model and a text model from the \emph{same} company are thus less alike in the senses they choose than two image models from \emph{different} companies. This complements the model-specific signatures that make generators individually identifiable~\citep{cekinmez2026guess}: in the space of which senses a model picks, what it generated matters more than who made it.
\newpage
\subsection{Investigating Potential Sources of Sense Collapse}
\label{sec:collapse}

\begin{wraptable}{r}{0.5\linewidth}
\centering
\small
\vspace{-\baselineskip}
\begin{tabular}{llccc}
\toprule
Model & Tuning & Base & Tuned & $\Delta$ \\
\midrule
SDXL     & DPO & 0.25 & 0.18 & $-0.07$ \\
SimpleAR & RL  & 0.11 & 0.10 & $-0.01$ \\
\bottomrule
\end{tabular}
\caption{Normalized entropy ($H_{\mathrm{norm}}$) over real senses, before and after preference tuning. DPO sharply reduces SDXL's diversity; SimpleAR's RL barely moves it.}
\label{tab:rl}
\end{wraptable}
\textbf{Preference tuning narrows senses, but does not fully explain the collapse.} A candidate driver is the preference optimization applied during alignment, which rewards a single preferred output and can erode the spread of a model's distribution. We test this with base-versus-tuned ablations on two open image generators, SDXL~\citep{podell2023sdxl} and SimpleAR~\citep{wang2025simplear}, each evaluated over the same 100 words. The effect is clear for SDXL: Diffusion-DPO~\citep{wallace2024diffusion}, a diffusion adaptation of direct preference optimization~\citep{rafailov2023direct}, lowers its normalized entropy over real senses from $0.25$ to $0.18$ ($\Delta = -0.07$), a genuine concentration onto fewer senses (Table~\ref{tab:rl}). For SimpleAR it is weak: reinforcement learning with GRPO~\citep{shao2024deepseekmath} leaves sense-entropy almost unchanged ($0.11 \rightarrow 0.10$); what it reduces is the share of \emph{unclassifiable} outputs ($23\% \rightarrow 19\%$) rather than the spread over real senses. Preference tuning thus contributes to the narrowing but cannot, on its own, account for the full modality gap.

\section{Conclusion}

When a word can mean several things, which meaning(s) does a model choose, and does that choice look the same textually and visually? By stripping 100 polysemous words of context and presenting the bare word to text-to-image and text-generation models, we turned an everyday property of language into a measurable distribution over senses, directly comparable across modalities and against how people respond to the same words.

Our central finding is a multimodal gap. Image models collapse onto a far narrower set of meanings than text models in the same family, and both fall well short of the diversity people show when they imagine or use these words. This narrowing is not a failure of knowledge. When asked, the models describe a word's meanings as broadly distributed. The gap appears only at generation, and it is largely the same across English, Turkish, and French, pointing to a property of how these systems interpret meaning rather than an artifact of one language or prompt.

Viewed together, our results suggest that representing ambiguity and expressing it are distinct challenges that are shaped by common finetuning practices such as RLHF. A model can know that ``bank'' has many senses and still, repeatedly draw only the riverbank when prompted. As multimodal systems increasingly shape what people read, see, and imagine, addressing this narrowing of meaning becomes a concrete step towards building AI systems whose grasp of meaning transfers honestly across modalities.

\bibliography{colm2026_conference}
\bibliographystyle{colm2026_conference}
\newpage
\appendix

\section{Model Inventory}
\label{app:models}
The complete set of evaluated models. The central panel (below) is used throughout; two open comparison models with published training recipes are used only for the targeted analyses.

\begin{small}
\begin{longtable}{lll}
\toprule
\textbf{Provider} & \textbf{Model} & \textbf{Reference} \\
\midrule
\endhead
\rowcolor{gray!12}\multicolumn{3}{l}{\textbf{Image generation}} \\
Black Forest Labs & FLUX.1.1 Pro & \citep{bfl2024flux11} \\
Black Forest Labs & FLUX.2 Flex & \citep{bfl2025flux2} \\
Black Forest Labs & FLUX.2 Pro & \citep{bfl2025flux2} \\
Black Forest Labs & FLUX.2 Max & \citep{bfl2025flux2max} \\
Google & Gemini 2.5 Flash & \citep{gemini2025flashimage25} \\
Google & Gemini 3.1 Flash & \citep{gemini2026flashimage31} \\
Google & Gemini 3 Pro & \citep{gemini2025proimage3} \\
xAI & Grok Base & \citep{xai2026grokimagine} \\
xAI & Grok Pro & \citep{xai2026grokimaginequality} \\
xAI & Grok Quality & \citep{xai2026grokimaginequality} \\
OpenAI & GPT Image 1 & \citep{openai2025gptimage1} \\
OpenAI & GPT Image 1 Mini & \citep{openai2025gptimagemini} \\
OpenAI & GPT Image 1.5 & \citep{openai2025gptimage15} \\
OpenAI & GPT Image 2 & \citep{openai2026gptimage2} \\
Alibaba & Qwen-Image & \citep{wu2025qwenimage} \\
Alibaba & Qwen-Image 2.0 & \citep{wu2026qwenimage2} \\
Alibaba & Qwen-Image 2.0 Pro & \citep{wu2026qwenimage2} \\
Z.ai & GLM Image & \citep{zai2025glmimage} \\
\midrule
\rowcolor{gray!12}\multicolumn{3}{l}{\textbf{Text generation}} \\
Google & Gemini 2.5 Flash & \citep{geminiteam2025gemini25} \\
Google & Gemini 3.5 Flash & \citep{gemini2026flash35} \\
Google & Gemini 3.1 Pro & \citep{gemini2026pro31} \\
Z.ai & GLM-5.1 & \citep{glm2026glm51} \\
Z.ai & GLM-5.2 & \citep{glm2026glm52} \\
xAI & Grok 4.20 & \citep{xai2026grok420} \\
xAI & Grok 4.20 NR & \citep{xai2026grok420} \\
xAI & Grok 4.3 & \citep{xai2026grok43} \\
OpenAI & GPT-3.5 Turbo & \citep{ouyang2022training} \\
OpenAI & GPT-4o & \citep{openai2024gpt4ocard} \\
OpenAI & GPT-5 & \citep{openai2025gpt5card} \\
OpenAI & GPT-5.4 Nano & \citep{openai2026gpt54} \\
OpenAI & GPT-5.4 Mini & \citep{openai2026gpt54} \\
OpenAI & GPT-5.5 & \citep{openai2026gpt55} \\
Alibaba & Qwen2.5 7B & \citep{qwen2024qwen25} \\
Alibaba & Qwen3.5 9B & \citep{qwen2026qwen35} \\
Alibaba & Qwen3.5 397B & \citep{qwen2026qwen35} \\
\bottomrule
\end{longtable}
\end{small}

\paragraph{Comparison models.} The preference-tuning ablation (\S\ref{sec:collapse}) holds the base model fixed and compares:
\begin{center}\small
\begin{tabular}{ll}
\toprule
Checkpoint & Reference \\
\midrule
SDXL 1.0, base & \citet{podell2023sdxl} \\
SDXL 1.0 ${+}$ Diffusion-DPO & \citet{wallace2024diffusion} \\
SimpleAR, SFT checkpoint & \citet{wang2025simplear} \\
SimpleAR ${+}$ GRPO (RL) & \citet{wang2025simplear, shao2024deepseekmath} \\
\bottomrule
\end{tabular}
\end{center}
The architecture analysis reuses the image panel above, each model labeled diffusion or autoregressive.

\section{Prompts}
\label{app:prompts}

Below, \texttt{\{word\}} is the stimulus and the candidate senses are the word's entries from Appendix~\ref{app:words}.

\paragraph{Generation.} The image prompt is the bare word (language-agnostic):
\begin{promptbox}\upshape\ttfamily \{word\}\end{promptbox}
Text models are instructed in the word's own language. English:
\begin{promptbox}Use the following word in a single sentence: \{word\}. Reply with only the sentence and nothing else.\end{promptbox}
Turkish:
\begin{promptbox}A\c{s}a\u{g}{\i}daki kelimeyi tek bir c\"umlede kullan{\i}n: \{word\}. Yaln{\i}zca c\"umleyi yaz{\i}n, ba\c{s}ka hi\c{c}bir \c{s}ey yazmay{\i}n.\end{promptbox}
French:
\begin{promptbox}Utilisez le mot suivant dans une seule phrase~: \{word\}. R\'epondez uniquement avec la phrase et rien d'autre.\end{promptbox}
The image-framing (``imagine'') control used in the stated-versus-revealed analysis instead asks:
\begin{promptbox}What image comes to mind when you think of the word ``\{word\}''? Describe it briefly.\end{promptbox}

\paragraph{Sense classification (judges).} Each output is labeled by both judges (GPT-5.4 and Gemini-3.5-Flash). The image judge:
\begin{promptbox}A text-to-image model was given only the single word ``\{word\}'' as its prompt, with no other context. Look at the image and decide which meaning of ``\{word\}'' it depicts. Choose exactly one of these labels: \upshape[the word's candidate senses]\itshape, plus \upshape\texttt{multiple}\itshape{} (two or more senses at once), \upshape\texttt{unclear}\itshape, and \upshape\texttt{other}\itshape. Respond with only the label.\end{promptbox}
The text judge is identical except it shows the produced sentence in place of the image:
\begin{promptbox}A language model was given only the single word ``\{word\}'' as its prompt, with no other context, and asked to use it in one sentence. Here is the sentence it produced: ``\upshape[sentence]\itshape''. Decide which meaning of ``\{word\}'' the sentence uses. Choose exactly one of these labels: \upshape[candidate senses]\itshape, plus \upshape\texttt{multiple}\itshape, \upshape\texttt{unclear}\itshape, \upshape\texttt{other}\itshape. Respond with only the label.\end{promptbox}
A \texttt{multiple} verdict triggers a follow-up:
\begin{promptbox}This output clearly uses two or more distinct meanings of ``\{word\}'' together. Of the meanings below, list every one that is actually present: \upshape[candidate senses]\itshape. Return the senses that apply (at least two).\end{promptbox}

\paragraph{Sense-distribution elicitation.} For the stated-versus-revealed comparison (Table~\ref{tab:stated_revealed}), each text model estimates how a word's senses are distributed. \emph{Predicting people}, sentence framing:
\begin{promptbox}Consider the word ``\{word\}''. Its possible meanings are listed below. If 100 different people were each asked to use this word in a single sentence, what percentage would use each meaning? Give a percentage for every meaning; they should sum to 100. \upshape[candidate senses]\end{promptbox}
\emph{Predicting people}, image framing:
\begin{promptbox}\ldots{} If 100 different people were each asked what image the word brings to mind, what percentage would picture each meaning? \ldots\end{promptbox}
\emph{Predicting itself}, sentence framing:
\begin{promptbox}\ldots{} If you were asked to use this word in a single sentence 100 times, what percentage of your sentences would use each meaning? \ldots\end{promptbox}
\emph{Predicting itself}, image-generation framing:
\begin{promptbox}\ldots{} If you were asked to generate an image from this word 100 times, what percentage of your images would depict each meaning? \ldots\end{promptbox}
Each model answers with one percentage per sense, which we normalize to a distribution.

\paragraph{Human baseline.} Prolific participants saw the bare word under two framings matched to the modalities; their responses are scored by the same judges and label set, with ``a person'' in place of ``a model''. The exact participant instructions are given in Appendix~\ref{app:human}.

\section{Words and Senses}
\label{app:words}
The complete stimulus set: every polysemous word and its candidate senses, for each language.

\subsection{English (100 words)}
\begin{longtable}{p{0.18\linewidth} p{0.74\linewidth}}
\toprule
\textbf{Word} & \textbf{Senses} \\
\midrule
\endhead
\texttt{trunk} & tree, car, elephant, body, luggage \\
\texttt{jordan} & country, person, shoes \\
\texttt{bolt} & fastener, lightning, run, lock \\
\texttt{bank} & finance, river, tilt, heap \\
\texttt{pitch} & throw, tone, sales, field \\
\texttt{bow} & bend, weapon, knot, ship, violin \\
\texttt{spring} & season, coil, water, leap \\
\texttt{crane} & bird, machine, stretch \\
\texttt{club} & organization, venue, weapon, golf, cards \\
\texttt{turkey} & bird, country, food, bowling, failure \\
\texttt{shot} & gunfire, injection, attempt, drink, photo \\
\texttt{bar} & pub, rod, block, law, music \\
\texttt{mercury} & planet, element, god \\
\texttt{jack} & lift, cards, name, connector \\
\texttt{mole} & animal, skin, spy, chemistry, sauce \\
\texttt{charge} & electric, fee, rush, accusation, responsibility \\
\texttt{key} & lock, crucial, music, keyboard, legend \\
\texttt{watch} & timepiece, observe, guard \\
\texttt{ring} & jewelry, sound, circle, arena, call \\
\texttt{nail} & finger, fastener, succeed \\
\texttt{chip} & fragment, microchip, snack, token \\
\texttt{cell} & biology, prison, phone, battery, spreadsheet \\
\texttt{toast} & bread, tribute, doomed \\
\texttt{track} & path, railroad, song, follow, athletics \\
\texttt{wave} & water, gesture, physics, surge \\
\texttt{train} & railway, teach, gown, sequence \\
\texttt{anchor} & ship, news, secure \\
\texttt{ash} & residue, tree \\
\texttt{ball} & sphere, dance, fun \\
\texttt{band} & music, strip, unite, frequency \\
\texttt{bark} & dog, tree, ship \\
\texttt{barrel} & cask, gun, oil \\
\texttt{basket} & container, basketball \\
\texttt{bass} & sound, fish, instrument \\
\texttt{batter} & baseball, cooking, beat \\
\texttt{beam} & light, support, smile, transmit \\
\texttt{bench} & seat, sports, court, workbench \\
\texttt{block} & solid, obstruct, city \\
\texttt{bowl} & dish, roll, stadium \\
\texttt{box} & container, fight, rectangle \\
\texttt{brush} & tool, touch, shrubs \\
\texttt{buck} & deer, dollar, resist \\
\texttt{button} & fastener, press, badge \\
\texttt{cane} & stick, plant, beat \\
\texttt{cape} & cloak, land \\
\texttt{capital} & city, money, letter, punishment \\
\texttt{case} & container, legal, instance, investigation \\
\texttt{chest} & body, box \\
\texttt{coach} & trainer, vehicle, tutor, class \\
\texttt{cobbler} & shoemaker, dessert \\
\texttt{apple} & fruit, company \\
\texttt{amazon} & river, rainforest, company, warrior \\
\texttt{mars} & planet, god, candy \\
\texttt{jaguar} & animal, car \\
\texttt{mustang} & horse, car \\
\texttt{cobra} & snake, yoga \\
\texttt{python} & snake, language \\
\texttt{fox} & animal, cunning, network \\
\texttt{cardinal} & bird, clergy, main, number \\
\texttt{ram} & sheep, force, memory \\
\texttt{paddle} & oar, bat, wade \\
\texttt{staple} & fastener, essential \\
\texttt{pitcher} & baseball, jug \\
\texttt{shuttle} & transport, spacecraft, badminton, weaving \\
\texttt{tap} & faucet, hit, dance, access \\
\texttt{drill} & tool, exercise, military \\
\texttt{pump} & device, shoe, inflate \\
\texttt{socket} & outlet, joint, tool \\
\texttt{switch} & control, change, rod \\
\texttt{iron} & metal, appliance, press, golf \\
\texttt{crown} & headwear, top, monarchy \\
\texttt{scale} & weigh, skin, music, size, climb \\
\texttt{plane} & aircraft, surface, tool \\
\texttt{fan} & cooling, admirer, spread \\
\texttt{bridge} & crossing, card game, nose, ship \\
\texttt{port} & harbor, left, wine, connector \\
\texttt{lodge} & cabin, file, stuck \\
\texttt{mine} & excavation, explosive, possessive \\
\texttt{grave} & burial, serious \\
\texttt{cast} & actors, throw, medical, mold \\
\texttt{press} & media, push, printing \\
\texttt{post} & mail, pole, job, publish, station \\
\texttt{tip} & gratuity, point, advice, tilt \\
\texttt{will} & volition, testament, future \\
\texttt{file} & document, submit, tool, line \\
\texttt{mortar} & cement, bowl, weapon \\
\texttt{hide} & conceal, skin \\
\texttt{tank} & container, vehicle, fail \\
\texttt{seal} & animal, close, stamp \\
\texttt{strike} & hit, labor, baseball, bowling \\
\texttt{stock} & shares, supply, broth, livestock \\
\texttt{bug} & insect, defect, annoy, surveillance \\
\texttt{fly} & insect, travel, zipper \\
\texttt{bat} & animal, sports, hit \\
\texttt{pen} & writing, enclosure, prison \\
\texttt{pool} & swimming, billiards, shared, puddle \\
\texttt{rock} & stone, music, sway \\
\texttt{fire} & flames, dismiss, shoot \\
\texttt{match} & firestick, contest, pairing \\
\texttt{tie} & necktie, fasten, draw, bond \\
\bottomrule
\end{longtable}

\subsection{Turkish (25 words)}
\begin{longtable}{p{0.18\linewidth} p{0.74\linewidth}}
\toprule
\textbf{Word} & \textbf{Senses} \\
\midrule
\endhead
\texttt{yüz} & face, hundred, swim, surface \\
\texttt{yaz} & summer, write \\
\texttt{gül} & rose, laugh \\
\texttt{dolu} & full, hail \\
\texttt{ay} & moon, month \\
\texttt{kaz} & goose, dig \\
\texttt{bağ} & vineyard, bond, tie \\
\texttt{kol} & arm, sleeve, branch \\
\texttt{baş} & head, beginning, chief \\
\texttt{ocak} & january, stove, quarry \\
\texttt{koy} & bay, put \\
\texttt{fiş} & plug, receipt, chip \\
\texttt{halka} & ring, public \\
\texttt{yıldız} & celestial, celebrity, symbol \\
\texttt{aç} & hungry, open \\
\texttt{saç} & hair, scatter, sheet \\
\texttt{sağ} & right, alive \\
\texttt{diz} & knee, lineup \\
\texttt{çalmak} & steal, play, ring \\
\texttt{boğaz} & throat, strait \\
\texttt{çene} & chin, chatter \\
\texttt{kök} & root, origin, math \\
\texttt{dal} & branch, dive, field \\
\texttt{acı} & pain, spicy, bitter \\
\texttt{kaymak} & slip, cream \\
\bottomrule
\end{longtable}

\subsection{French (25 words)}
\begin{longtable}{p{0.18\linewidth} p{0.74\linewidth}}
\toprule
\textbf{Word} & \textbf{Senses} \\
\midrule
\endhead
\texttt{trombone} & instrument, paperclip \\
\texttt{souris} & animal, computer \\
\texttt{timbre} & stamp, tone, buzzer \\
\texttt{chaîne} & chain, channel, range, assembly \\
\texttt{avocat} & lawyer, avocado \\
\texttt{pile} & battery, stack, heads, exactly \\
\texttt{clé} & key, wrench, clef \\
\texttt{bouchon} & cork, jam, bistro \\
\texttt{règle} & ruler, rule, period \\
\texttt{cadre} & frame, context, executive \\
\texttt{fève} & bean, trinket \\
\texttt{mine} & expression, mine, lead, landmine \\
\texttt{toile} & canvas, cloth, web, internet \\
\texttt{prise} & outlet, grip, catch, dose \\
\texttt{pont} & bridge, deck, weekend \\
\texttt{coup} & blow, attempt, coup \\
\texttt{batterie} & drums, battery, cookware \\
\texttt{filet} & net, fillet, trickle \\
\texttt{fort} & strong, loud, fort, skilled \\
\texttt{gomme} & eraser, gum, rubber \\
\texttt{planche} & plank, plate, exercise \\
\texttt{canon} & cannon, standard, barrel, gorgeous \\
\texttt{pouce} & thumb, inch, truce \\
\texttt{chouette} & owl, cool \\
\texttt{vase} & vase, mud \\
\bottomrule
\end{longtable}

\section{Per-Model Diversity and Fallback Rates}
\label{app:fallback}
Normalized entropy over real senses ($H_{\mathrm{norm}}$, fallback excluded) and the fraction of outputs assigned each fallback label, \texttt{multiple} (two or more senses at once), \texttt{unclear}, \texttt{other}, and their total, per model and for the human baselines. Image models carry more fallback mass than text models; notably the \texttt{unclear} share is small everywhere, and humans' fallback is dominated by \texttt{multiple} (people picturing several senses) rather than \texttt{unclear}.

\begin{small}
\setlength{\tabcolsep}{4pt}
\begin{longtable}{lccccc}
\toprule
Model & $H_{\mathrm{norm}}$ & Mult.\% & Uncl.\% & Other\% & All\% \\
\midrule
\endhead
\multicolumn{6}{l}{\textit{Image generation}} \\
GPT Image 1 & 0.075 & 0.3 & 1.4 & 2.5 & 4.2 \\
GPT Image 1 Mini & 0.058 & 0.4 & 1.4 & 2.8 & 4.5 \\
GPT Image 1.5 & 0.063 & 0.4 & 2.5 & 4.4 & 7.3 \\
GPT Image 2 & 0.050 & 0.1 & 2.4 & 2.2 & 4.7 \\
Grok Base & 0.045 & 3.0 & 0.1 & 0.8 & 4.0 \\
Grok Pro & 0.143 & 1.1 & 0.4 & 1.0 & 2.6 \\
Grok Quality & 0.081 & 1.2 & 0.3 & 1.0 & 2.5 \\
Gemini 2.5 Flash & 0.132 & 0.8 & 0.3 & 1.9 & 3.1 \\
Gemini 3.1 Flash & 0.161 & 1.2 & 0.0 & 1.6 & 2.8 \\
Gemini 3 Pro & 0.142 & 3.4 & 0.1 & 0.8 & 4.4 \\
Qwen-Image & 0.070 & 0.0 & 7.4 & 9.8 & 17.2 \\
Qwen-Image 2.0 & 0.133 & 0.2 & 15.3 & 14.4 & 29.9 \\
Qwen-Image 2.0 Pro & 0.098 & 0.0 & 3.5 & 5.3 & 8.9 \\
FLUX.1.1 Pro & 0.129 & 0.6 & 4.1 & 5.8 & 10.4 \\
FLUX.2 Flex & 0.113 & 8.0 & 0.2 & 1.0 & 9.2 \\
FLUX.2 Pro & 0.130 & 7.4 & 0.4 & 0.9 & 8.7 \\
FLUX.2 Max & 0.136 & 10.5 & 0.3 & 0.7 & 11.5 \\
\midrule \multicolumn{6}{l}{\textit{Text generation}} \\
GPT-3.5 Turbo & 0.266 & 0.5 & 0.5 & 3.0 & 4.0 \\
GPT-4o & 0.297 & 0.7 & 0.2 & 3.0 & 3.9 \\
GPT-5 & 0.323 & 0.4 & 0.0 & 3.2 & 3.7 \\
GPT-5.4 Nano & 0.166 & 0.3 & 0.7 & 4.6 & 5.6 \\
GPT-5.4 Mini & 0.219 & 0.6 & 0.7 & 4.6 & 5.8 \\
GPT-5.5 & 0.136 & 0.1 & 0.6 & 2.1 & 2.8 \\
Grok 4.20 & 0.363 & 0.4 & 0.4 & 6.1 & 6.9 \\
Grok 4.20 NR & 0.272 & 0.2 & 0.7 & 5.2 & 6.1 \\
Grok 4.3 & 0.286 & 0.2 & 0.0 & 2.6 & 2.8 \\
Gemini 2.5 Flash & 0.330 & 0.6 & 0.0 & 5.6 & 6.2 \\
Gemini 3.5 Flash & 0.173 & 0.0 & 0.0 & 2.8 & 2.8 \\
Gemini 3.1 Pro & 0.210 & 0.0 & 0.0 & 3.7 & 3.7 \\
Qwen2.5 7B & 0.212 & 0.6 & 1.3 & 5.7 & 7.6 \\
Qwen3.5 9B & 0.217 & 0.0 & 0.2 & 3.1 & 3.2 \\
Qwen3.5 397B & 0.212 & 0.0 & 0.0 & 2.9 & 2.9 \\
GLM-5.1 & 0.274 & 0.0 & 0.0 & 3.8 & 3.8 \\
GLM-5.2 & 0.286 & 0.0 & 0.0 & 3.3 & 3.3 \\
\midrule \multicolumn{6}{l}{\textit{Humans}} \\
Humans (image) & 0.473 & 8.7 & 1.2 & 4.5 & 14.4 \\
Humans (sentence) & 0.496 & 2.5 & 0.9 & 5.5 & 8.9 \\
\bottomrule
\end{longtable}
\end{small}

\section{Part-of-Speech Robustness}
\label{app:pos}
Our word panel is noun-heavy, which raises the concern that the multimodal gap might be driven by words that afford a salient verb reading (e.g., \emph{bolt}, \emph{charge}, \emph{spring}), for which the image and text modalities could behave differently. To rule this out, we split the 100 English words into two halves: the 50 that carry a clear non-noun sense---45 with a common verb sense, three adjectives (\emph{key}, \emph{cardinal}, \emph{grave}), one pronoun (\emph{mine}), and one auxiliary (\emph{will})---and the 50 that are noun-dominant. The split is defined at the level of the word, i.e., whether the form affords a non-noun reading at all, not the sense that any individual sample happened to realize.

The two halves are statistically indistinguishable in sense diversity within each modality. Image models reach $H_\textrm{norm}=0.13$ on both halves (Mann--Whitney $U$-test, $p=0.52$, Cohen's $d=-0.06$), and text models reach $0.27$ on the verb-capable half versus $0.25$ on the noun-dominant half ($p=0.52$, $d=+0.12$). If anything, the direction runs opposite to the concern: verb-capable words are marginally \emph{more} diverse in text, not less. Crucially, the multimodal gap holds within each half: image entropy falls well below text entropy for the verb-capable words ($0.13$ vs.\ $0.27$; paired Wilcoxon $p < 10^{-4}$) and for the noun-dominant words ($0.14$ vs.\ $0.25$; $p < 10^{-5}$) alike. The gap is therefore a property of the two modalities, not of the grammatical class of the probe words.

\section{Ordering versus Calibration in Predicted Distributions}
\label{app:calibration}
When asked to predict the distribution of senses people assign to a word, models report distributions far more diverse than either human judgments or their own generations (Table~\ref{tab:stated_revealed}). This overestimation could arise in two distinct ways: the model may correctly identify which senses are common and merely assign them probabilities that are too even (a \emph{calibration} error), or it may misjudge which senses dominate in the first place (an \emph{ordering} error). We separate the two by comparing each of the 15 text models' predicted meaning distribution against the human meaning distribution for the same word, across $1{,}313$ model--word pairs.

The ordering is largely correct. Predicted and human distributions rank senses in close agreement (median Spearman $\rho = 0.87$, mean $0.70$; positive for $94\%$ of words, $p \approx 10^{-165}$), and the model's most-probable sense coincides with the human-dominant sense $72\%$ of the time, against a chance rate of $32\%$; the human-dominant sense falls within the model's top two predictions $92\%$ of the time. This holds for every model individually (all median $\rho \ge 0.67$, all top-1 concordance $\ge 0.63$).

What the models get wrong is the magnitude. Their predicted distributions are systematically flatter than reality, placing only $0.55$ of the probability mass on the dominant sense where humans place $0.72$, and inflating entropy by $+0.40$ bits. Because entropy is invariant to how the senses are ordered, this flattening---not any reshuffling of ranks---is by construction the entire source of the overdiversity. Consistent with this, the inflation persists even on the words where the model ranks the dominant sense correctly ($+0.47$ bits), and is if anything larger there than on the words it ranks wrong ($+0.24$ bits). The same pattern holds when the reference is the model's own generated distribution rather than the human one (top-1 concordance $0.69$, median $\rho = 0.80$), where the flattening is larger still ($+0.81$ bits; mass on the dominant sense $0.53$ versus $0.86$). Models thus know which senses are common and which are rare; they overestimate diversity by assigning those senses probabilities that are too uniform, not by mistaking which sense dominates.

\section{Human Study}
\label{app:human}
\paragraph{Recruitment.} We collected the human baseline on Prolific over the same 100 English words. In total, 540 participants completed the study in June 2026, recruited as fluent English speakers with a minimum approval rate of $99\%$ and at least 50 prior Prolific submissions. We placed no restriction on country of residence.

\paragraph{Task.} Each participant responded to five polysemous words. In the meaning condition they wrote a sentence, \textit{``Use the word `\{word\}' in a sentence that clearly shows its meaning. Avoid vague sentences that could fit any word.''}; in the image condition they described the image the word brings to mind, \textit{``What image comes to mind when you think of the word `\{word\}'? Describe it briefly.''} Participants spent a median of $34$ seconds per word. This yielded $2{,}782$ responses ($1{,}393$ sentence, $1{,}389$ image), about $14$ per word per condition.

\paragraph{Scoring.} Responses were classified into each word's candidate senses by the same judge (GPT-5.4) used for the models, with the prompt referring to ``a person'' rather than ``a model'' (Appendix~\ref{app:prompts}).

\end{document}